\documentclass[conference]{IEEEtran}

\usepackage[tight,footnotesize]{subfigure}
\usepackage{float}
\usepackage{url}
\usepackage{latexsym}
\usepackage{epstopdf}
\usepackage[dvipdfm]{graphicx}
\usepackage[usenames,dvipsnames]{color}
\usepackage{amsmath}
\usepackage{amsfonts}
\usepackage{amssymb}
\usepackage{graphicx}
\usepackage{multicol}
\usepackage{verbatim}
\usepackage{enumerate}
\usepackage{cite}
\usepackage{array}
\usepackage{algorithm}
\usepackage{algorithmic}

\begin{document}

\title{Distributed Cooperative Q-learning for Power Allocation in Cognitive Femtocell Networks}


\author{Hussein Saad\IEEEauthorrefmark{1}, Amr Mohamed\IEEEauthorrefmark{2} and Tamer ElBatt\IEEEauthorrefmark{1}\\
\normalsize \begin{tabular}{cc}
\IEEEauthorrefmark{1}Wireless Intelligence Network Center (WINC), & \hspace{.5in} \IEEEauthorrefmark{2}Computer Science and Engineering Department\\
Nile University, Cairo, Egypt. & \hspace{.5in} Qatar University, P.O. Box 2713, Doha, Qatar.\\
hussein.saad@nileu.edu.eg &  \hspace{.5in} amrmg@qu.edu.qa\\
telbatt@nileuniversity.edu.eg \end{tabular}
}



%


\maketitle


\begin{abstract}
In this paper, we propose a distributed reinforcement learning (RL) technique called distributed power control using Q-learning (DPC-Q) to manage the interference caused by the femtocells on macro-users in the downlink. The DPC-Q leverages Q-Learning to identify the sub-optimal pattern of power allocation, which strives to maximize femtocell capacity, while guaranteeing macrocell capacity level in an underlay cognitive setting. We propose two different approaches for the DPC-Q algorithm: namely, independent, and cooperative. In the former, femtocells learn independently from each other while in the latter, femtocells share some information during learning in order to enhance their performance. Simulation results show that the independent approach is capable of mitigating the interference generated by the femtocells on macro-users. Moreover, the results show that cooperation enhances the performance of the femtocells in terms of speed of convergence, fairness and aggregate femtocell capacity.
\end{abstract}


%
\IEEEpeerreviewmaketitle

\section{Introduction}\label{intro}

Femtocells are considered to be a highly promising solution for the enhancement of the indoor coverage problem. However, femtocells are deployed unpredictably in the macrocell area. Thus, their interference on macro-users and other femtocells is considered to be a daunting problem \cite{4623708}, \cite{femto}.


Since femtocells are installed by the end user, their number and positions are random and unknown to the network operator. This makes the centralized approach for solving the interference problem very hard due to the huge overhead needed which in turn calls for a distributed interference management strategy. In the distributed scheme, each femtocell needs to \textbf{\emph{learn}} how to interact with the dynamic environment created by the coexistence of the femto and macro cells in order to adjust its parameters (carrier frequency and transmission power) to satisfy the QoS of its own users while guaranteeing certain QoS for the macrocell users.

Based on these observations, in this paper we focus on closed access femtocells \cite{closed} working in the same bandwidth with macrocells (cognitive femtocells). We will use a distributed machine learning technique called reinforcement learning (RL) \cite{introtoRL} to handle the interference problem generated by the femtocells on the macrocells' users. One of the most popular RL techniques is Q-learning \cite{q-learning}. The reason we chose Q-learning is because it finds optimal decision policies without any prior model of the environment (in our settings, a prior model can not be achieved due to the unplanned placement of the femtocells). Moreover, Q-learning allows the agents (i.e the femtocells) to take actions while they are learning (i.e no need for a centralized approach). These features make Q-learning very suitable to be applied to the distributed femtocell setting in the form of the so called multi-agent Q-learning (\textbf{MAQL}) \cite{collaborativeRL}. In this paper, MAQL is applied in two different paradigms: independent learning (\textbf{IL}) and cooperative learning (\textbf{CL}). The former assumes that agents are unaware of the other agents' actions while the latter allows the agents to share some knowledge while they are learning to enhance their performance\cite{expertQ,collaborativeRL}.

In literature, RL has been used to perform power allocation in wireless networks. In \cite{3},\cite{4}, authors used IL Q-learning to perform power allocation in order to control the aggregate interference generated by multiple secondary users on the primary receiver of a digital TV (DTV) system. In \cite{1}, authors addressed the same goal of interference control but in the context of OFDMA-based femtocells. In \cite{2}, authors used IL Q-learning in the context of cognitive femtocells and introduced a new concept called docitive femtocells. However, all the papers discussed above were interested in maintaining the QoS of the primary users and ignored the QoS of the femtocells (e.g: fairness, maximizing the femtocell capacity). Moreover, they all used the IL paradigm and did not take into consideration any cooperation between the agents (femtocells) during the learning process.

Motivated by this, in this paper we apply Q-learning for power control in closed access cognitive femtocells network. The contributions of this paper can be summed up as follows:

\begin{itemize}

\item A distributed algorithm based on IL paradigm is used to handle the interference problem. A new reward function is introduced and compared to the reward function used in literature  \cite{1},\cite{2}. The comparison is applied in two different scenarios:

    \begin{enumerate}
    \item Maintaining the QoS (i.e. the capacity) of the macrocell without taking into consideration the QoS of the femtocells.
    \item Enhancing the capacity of the femtocells while maintaining the QoS of the macrocell.
    \end{enumerate}
\item Cooperation between the femtocells is introduced to enhance the aggregate capacity and fairness amongst all the femtocells, while maintaining the macrocell QoS.

\end{itemize}

The remaining part of this paper is organized as follows. Section \ref{back} gives a brief background for the original single agent Q-learning. In section \ref{sys}, the system model is described. Section \ref{problem} introduces the proposed distributed Q-learning algorithm and the Q-learning formulation for the cognitive femtocells problem. The simulation scenario and the results are discussed in section \ref{scenario}. Finally the conclusion is given in section \ref{conclusion}.

\section{Background: Single Agent Q-learning (SAQL)}\label{back}

In this section, the idea of Q-learning is presented by introducing the single agent case \cite{q-learning}. The Q-learning model can be defined by the tuple $\lbrace S,A,R(s,a)\rbrace$ where $S=\lbrace s_1,s_2,\cdots,s_m\rbrace$ is the set of possible states the agent can occupy, $A=\lbrace a_1,a_2,\cdots,a_l\rbrace$ is the set of possible actions the agent can perform and $R(s,a)$ is the reward function that determines the reward fed back to the agent by the environment when performing action $a$ in state $s$. The interaction between the agent and the environment at time $t$ can be described as follows:

\begin{itemize}
\item The agent senses the environment and observes its current state $s_t \in S$.
\item Based on $s_t$, the agent selects action $a_t \in A$.
\item Based on $a_t$, the environment makes a transition to a new state $s_{t+1} \in S$ and as a result achieves a reward $r_t = R(s_t,a_t)$ due to this transition.
\item The reward is fed back to the agent and the process is repeated.
\end{itemize}

The end goal of the agent is to find an optimal policy $\pi^*(s)$, which defines the action to be selected for each state $s \in S$ in order to maximize the expected discounted reward over an infinite time:

\begin{equation}\label{eq.4}
V^{\pi}(s) = \mathbb{E}\lbrace\sum_{t=0}^{\infty}\gamma^t r(s_t,\pi(s))|s_o = s\rbrace
\end{equation}

where $V^{\pi}(s)$ is the value function of state $s$ which represents the expected discounted infinite reward when the initial state is $s_o$ and $0\leq \gamma \leq 1$ is the discount factor that determines how much effect future rewards have on the decisions at each moment. Furthermore, equation (\ref{eq.4}) can be expressed as \cite{1}:

\begin{equation}\label{eq.5}
V^{\pi}(s) = \mathbb{E}\lbrace r(s,\pi(s))\rbrace + \gamma \sum_{s^{'}\in S} P_{s,s^{'}}(\pi(s))V^{\pi}(s^{'})
\end{equation}

where $s^{'}$ is the new state to which the environment transits after taking action $a = \pi(s)$ and $P_{s,s^{'}}$ is the transition probability from state $s$ to state $s^{'}$ after performing action $a = \pi(s)$. From equation (\ref{eq.5}), the optimal value function $V^*(s)$ can be written as:

\begin{equation}\label{eq.6}
V^{*}(s) = \max_{a\in A} (\mathbb{E}\lbrace r(s,a)\rbrace + \gamma \sum_{s^{'}\in S}P_{s,s^{'}}(a)V^{*}(s^{'}))
\end{equation}

Q-learning aims at finding the optimal policy $\pi^{*}(s)$ that corresponds to $V^{*}(s)$ without having any prior knowledge about the transition probabilities $P_{s,s^{'}}$. In order to do this, a new value called Q-value is defined for each state-action pair, where the optimal Q-value is defined as:

\begin{equation}\label{eq.7}
Q^{*}(s,a) = \mathbb{E}\lbrace r(s,a)\rbrace + \gamma \sum_{s^{'}\in S}P_{s,s^{'}}(a)\max_{b\in A}Q^{*}(s^{'},b)
\end{equation}

Equation (\ref{eq.7}) states that the optimal value function can be expressed by $V^*(s) = \max_{a\in A} Q^{*}(s,a)$. Thus, if the optimal Q-value is known for each state-action pair, the optimal policy can be determined by $\pi^{*}(s) = arg\max_{a\in A}Q^{*}(s,a)$. The Q-learning algorithm finds $Q^{*}(s,a)$ in a recursive manner using a simple update rule:

\begin{equation}\label{eq.8}
Q(s,a) := (1-\alpha)Q(s,a) + \alpha(r(s,a) + \gamma \max_{b\in A}Q(s^{'},b))
\end{equation}

Where $\alpha$ is the learning rate. It was proved in \cite{q-learning}, \cite{qprove} that this update rule converges to the optimal Q-value under certain conditions. One of these conditions is that each state-action pair must be visited infinitely often \cite{q-learning}. To address this notion, a random number $\epsilon$ is introduced where at each step of the learning process the action is chosen according to $a = arg\max_{a\in A}Q(s,a)$ with probability $(1-\epsilon)$ or randomly with probability $\epsilon$. Moreover, in the convergence proof, the reward function is assumed to be bounded and deterministic for each state-action pair \cite{qprove}. However, in the multi-agent case, this condition is violated since the reward for each state will depend on the joint action of all agents, hence the reward function will not be deterministic from the agent point of view. Thus, in section \ref{scenario}, the effect of choosing the reward function will be addressed using simulations.

\section{System Model}\label{sys}

In this paper, a wireless network consisting of one macro cell with one single transmit and receive antenna denoted by Macro Base Station (MBS) underlaid with $N_{femto}$ femtocells each with one Femto Base Station (FBS) is considered. $U_m$ and $U_f$ macro and femto users are located randomly inside the macro and femto cells respectively. Both MBS and FBS's transmit over the same $N_{sub}$ sub-carriers where orthogonal downlink transmission is assumed in each time slot.

The transmission powers of the MBS and FBS $i$ in subcarrier $n$ are denoted by $P_{o}^{(n)}$ and $P_{i}^{(n)}$ respectively. Moreover, the maximum transmission powers for the MBS and FBS $i$ are $P_{max}^{m}$ and $P_{max}^{f}$ respectively, where $ \sum_{n=1}^{N_{sub}} P_{o}^{(n)} \leq P_{max}^{m} $ and $ \sum_{n=1}^{N_{sub}} P_{i}^{(n)} \leq P_{max}^{f} $.

The system performance is analyzed in terms of the capacity measured in (bits/sec/Hz). The capacity achieved by the MBS at its associated user in subcarrier $n$ is:

\begin{equation}
C_o^{(n)} = \log_{2}(1 + \frac{h_{oo}^{(n)} P_o^{(n)}}{\sum_{i=1}^{N_{femto}} h_{io}^{(n)}P_i^{(n)} + \sigma^{2}})
\end{equation}

where $h_{oo}^{(n)}$ denotes the channel gain between the MBS and its associated user in subcarrier $n$; $h_{io}^{(n)}$ denotes the channel gain between FBS $i$ and the macro user in subcarrier $n$ and $\sigma^2$ is the noise power. The capacity achieved by FBS $i$ at its associated user in subcarrier $n$ is:

\begin{equation}
C_i^{(n)} = \log_{2}(1 + \frac{h_{ii}^{(n)} P_i^{(n)}}{\sum_{j=1,j\neq i}^{N_{femto}} h_{ji}^{(n)}P_j^{(n)} + h_{oi}^{(n)}P_o^{(n)} + \sigma^{2}})
\end{equation}

where $h_{ii}^{(n)}$ denotes the channel gain between FBS $i$ and its associated user in subcarrier $n$; $h_{ji}^{(n)}$ denotes the channel gain between FBS $j$ and the femto user associated withe FBS $i$ in subcarrier $n$.

\section{Distributed Power Control using Q-learning (DPC-Q)}\label{problem}

In this section, a distributed MAQL technique called DPC-Q is presented where multiple agents (i.e: femtocells) aim at learning a sub-optimal decision policy (i.e: power allocation) by repeatedly interacting with the environment. First we describe the two different paradigms in which the proposed DPC-Q algorithm is applied: Independent learning (IL) and Cooperative learning (CL). Then, the agents, states, actions and reward functions used during the simulations will be introduced.

\begin{itemize}
\item \textbf{Independent learning (IL)}: In this paradigm, each agent learns independently from other agents (i.e: ignores other agents' actions and considers other agents as part of the environment). Although, this may lead to oscillations and convergence problems, the IL paradigm showed good results in many applications \cite{stateOfArt}. The action selection strategy for agent $i$ in the IL paradigm is the same as the SAQL case: $a_{i} = arg\max_{a\in A_{i}} Q_{i}(s_{i},a)$, where $A_{i}$ is the set of actions available for agent $i$ (in our settings, we assume that $A_{i}$ is the same for all agents $1 \leq i \leq N$, where $N$ is the number of agents). The only difference here compared to the SAQL case is that the reward function is now dependent on the joint action of all agents $\vec{a}$. Thus, the update rule can be rewritten as:

    \begin{equation}\label{eq.9}
    Q_{i}(s_{i},a_{i}):= (1-\alpha)Q_{i}(s_{i},a_{i})+ \alpha(r_{i}(s_{i},\vec{a}) +\gamma \max_{b\in A_{i}}Q_{i}(s_{i}^{'},b))
    \end{equation}
    
    However, in the multi-agent case, acting in an independent way is not always the best approach because agents now affect each other in terms of the reward function as shown in equation (\ref{eq.9}). So, agents will need to know some information about each other (e.g: states, action, Q-tables,$\cdots$,etc). This information is shared during the learning process in order to enhance the agents' performance. Motivated by this, we propose a mechanism where each agent shares a portion of its Q-table with all other agents \footnotemark (The Q-table is a table with $|S|$x$|A|$ entries where $|S|$ and $|A|$ are the total number of possible states and actions respectively).
    
    \footnotetext{We assume that the shared portion of the Q-table is put in the control bits of the packets transmitted between the femtocells. The details of the exact protocol lie out of the scope of this paper.}

\item \textbf{Cooperative learning (CL)}: CL is performed as follows:

    Agent $i$ shares the row of its Q-table that corresponds to its current state with all other agents $j$. Then agent $i$ selects its action according to the following equation:

    \begin{equation}\label{eq.10}
        a_i = arg\max_{a}(\sum_{1 \leq j \leq N}Q_{j}(s_{j},a))
    \end{equation}

    The main idea behind this strategy depends on two important observations: \textbf{$1)$} the meaning of the Q-value $Q(s,a)$, which is an estimate of the value of future rewards if the agent selects action $a$ in state $s$. For example, if the reward of a femtocell is its capacity, then at a certain instant, if the agent was in state $s$, has two actions $a1$, $a2$ and $Q(s,a1) > Q(s,a2)$, then choosing $a1$ in state $s$ would achieve higher femtocell capacity than $a2$. \textbf{$2)$} The definition of the global Q-value $\textbf{Q(s,a)}$, which represents the Q-value of the whole system (i.e. if the multi-agent scenario is transformed into a single agent one using a centralized controller with global state $\textbf{s}$ and global joint action $\textbf{a}$). This global Q-value can be decomposed into a linear combination of local agent-dependent Q-values: $\textbf{Q(s,a)} = \sum_{1 \leq j \leq N} Q_{j}(s_{j},a_{j})$ \cite{learningAndCooperating}. Thus, if each agent $j$ maximized its own Q-value, the global Q-value will be maximized. Based on these two observations, choosing the action based on equation \ref{eq.10} would maximize the global Q-value. However, the solution is still not global optimum because based on equation \ref{eq.10}, all agents will choose the same action. For example, if there are two agents (femtocells) $1$ and $2$, each agent has one state $s$ and three actions $a1$, $a2$ and $a3$, the reward for each agent is its capacity and the Q-values for both agents are as follows: $Q_{1}(s,a1) = 1$, $Q_{1}(s,a2) = 2$, $Q_{1}(s,a3) = 3$, $Q_{2}(s,a1) = 4$, $Q_{2}(s,a2) = 6$ and $Q_{2}(s,a3) = 4.5$, then in the IL paradigm, agent $1$ will choose action $a3$, thus maximizing its capacity, while agent $2$ will choose action $a2$, thus maximizing its capacity. However, in the CL paradigm, both agents will choose action $a2$ (the maximum of the summation of the Q-values is $2+6$), thus maximizing the aggregate capacity.

    In terms of overhead, according to our proposed cooperation algorithm each femtocell should only share a row of its Q-table with all its neighbors. This row has a size of $1$x$|A|$. So if the number of femtocells is $N_{femto}$, then the total overhead needed is $N_{femto}.(N_{femto}-1). |A|$ per unit time.

    Finally, it should be noticed that we assume that the information to be shared is put in the control bits in the packets transmitted between the femtocells. The different paradigms of the DPC-Q algorithm are summarized in algorithm \ref{algo.}.
\end{itemize}

\begin{algorithm}
\caption{The proposed DPC-Q algorithm}
\label{algo.}
\begin{algorithmic}
\STATE Let $t = 0$, $Q_{i}^{0}(s_i,a_i)=0$ for all $S_i \in A$ and $a_i \in A$
\STATE Initialize the starting state $s_{i}^{t}$
\STATE \textbf{loop}
\STATE send $Q_{i}^{t}(s_{i}^{t},:)$ to all other agents $j$
\STATE receive $Q_{j}^{t}(s_{j}^{t},:)$ from all other agents $j$
\IF {rand $< \epsilon$}
        \STATE select action randomly
\ELSE
        \IF {leaning paradigm == IL}
                \STATE choose action: $a_{i}^{t} = arg\max_{a} Q_{i}(s_{i}^{t},a)$
        \ELSE
                \STATE choose action: $a_{i}^{t} = arg\max_{a}(\sum_{1 \leq j \leq N}Q_{j}^{t}(s_{j}^{t},a))$
        \ENDIF
\ENDIF
\STATE receive reward $r_{i}^{t}$
\STATE observe next state $s_{i}^{t+1}$
\STATE update Q-table as in equation \ref{eq.9}
\STATE $s_{i}^{t} = s_{i}^{t+1}$
\STATE \textbf{end loop}
\end{algorithmic}
\end{algorithm}

The agents, states, actions and reward function are defined as follows:

\begin{itemize}
\item \textbf{Agent: }$FBS_i , \forall 1\leq i \leq N_{femto}$
\item \textbf{State: }At time instant $t$ for femtocell $i$ in subcarrier $n$, the state is defined as: $s_{t}^{i,n} = \lbrace I_{t}^{n},\emph{P}_{t}^{i} \rbrace$ where $I_{t}^{n} \in \lbrace 0,1 \rbrace$ indicates the level of interference measured at the macro-user in subcarrier $n$ at time $t$:

    \begin{equation}\label{eq.12}
    I_{t}^{n} = \begin{cases} 1,& C_{o}^{(n)} < \Gamma^{o} \\ 0,& C_{o}^{(n)} \geq \Gamma^{o} \end{cases}
    \end{equation}

    where $\Gamma^{o}$ is the target capacity determining the QoS performance of the macrocell. We assume that the macrocell reports the value of $C_{o}^{n}$ to all FBS through the backhaul connection.

    $\emph{P}_{t}^{i}$ determines the total power FBS $i$ is transmitting with at time $t$:

    \begin{equation}\label{eq.13}
    \emph{P}_{t}^{i} = \begin{cases} 0,& \sum_{n=0}^{N_{sub}} p_{t}^{i,n} < (P_{max}^{f} - A1) \\ 1,& (P_{max}^{f} - A2) \leq \sum_{n=0}^{N_{sub}} p_{t}^{i,n} \leq P_{max}^{f} \\ 2,& \sum_{n=0}^{N_{sub}} p_{t}^{i,n} > P_{max}^{f}\end{cases}
    \end{equation}

    where $P_{max}^{f}$, $A1$ and $A2$ are set to $15$, $5$ and $5$ dBm respectively in the simulations and $p_{t}^{i,n}$ is the power femtocell $i$ transmitting with on subcarrier $n$ at time $t$. \emph{It should be noticed that other values for $A1$ and $A2$ as well as more power levels were tried through the simulations and the performance gain was marginal.}
\item \textbf{Action: } The set of actions for each agent is the set of possible powers that the FBS can use. In the simulations a range from $-20$ to $15$ dBm with step of $2$ dBm is used.
\item \textbf{Reward: } Two different reward functions were considered in the simulations:

\begin{enumerate}
\item \begin{equation}\label{eq.14} r_{t}^{i,n} = \begin{cases} e^{(-(C_{o}^{(n)} - \Gamma^{o})^2)},& \sum_{n=0}^{N_{sub}} p_{t}^{i,n} \leq P_{max}^{f} \\ -1,&   \sum_{n=0}^{N_{sub}} p_{t}^{i,n} > P_{max}^{f}\end{cases}\end{equation}

    The rationale behind this reward function is to maintain the capacity of the macrocell at the target capacity $\Gamma^{o}$ while not exceeding the allowed $P_{max}^{f}$. The reason for the small difference between the positive (when $P_{max}^{f}$ is not exceeded) and negative (when $P_{max}^{f}$ is exceeded) rewards is due to the way the states are defined. Since the state $s_{t}^{i,n}$ is defined as $\lbrace I_{t}^{n}, \emph{P}_{t}^{i} \rbrace$ and $\emph{P}_{t}^{i}$ is defined for certain ranges of powers not for discrete power levels, therefore, large negative numbers can not be assigned as a reward when $P_{max}^{f}$ is exceeded. For example, if $I_{t}^{n} = 1$ and $\emph{P}_{t}^{i} = 6$ dBm, then FBS $i$ is in state $\lbrace 1,0 \rbrace$ in subcarrier $n$. If FBS $i$ took the action $a_{t}^{i,n} = 8$ dBm, then the next state would be $\lbrace 1,1 \rbrace$ and FBS $i$ is rewarded positively according to equation \ref{eq.14}. Now consider the case when $I_{t}^{n} = 1$ and $\emph{P}_{t}^{i} = 9$ dBm, then FBS $i$ is again in state $\lbrace 1,0 \rbrace$ in subcarrier $n$. If FBS $i$ took the same action $a_{t}^{i,n} = 8$ dBm, then the next state would $\lbrace 1,2 \rbrace$ and FBS $i$ is rewarded $-1$. So from this example, it can be shown that different rewards could be assigned for the same state-action pair. Thus, the difference between these different rewards must not be large. If the state was defined for discrete power levels (e.g: $\emph{P}_{t}^{i} = \sum_{n=1}^{N_{sub}}p_{t}^{i,n}$), then it would be possible to assign rewards with large differences because the case of having different rewards for the same state-action pair will not occur. However, defining the states in a discrete manner would dramatically increase the number of possible states which in turn makes it harder to satisfy the condition of visiting all the state-action pairs infinitely many times. Based on this observation, in the next section we compare our reward function to the reward function used in \cite{1}:

    \begin{equation}\label{eq.15} r_{t}^{i,n} = \begin{cases} K -(C_{o}^{(n)} - \Gamma^{o})^2,& \sum_{n=0}^{N_{sub}} p_{t}^{i,n} \leq P_{max}^{f} \\ 0,&   \sum_{n=0}^{N_{sub}} p_{t}^{i,n} > P_{max}^{f}\end{cases}\end{equation}

    where $K$ is a constant value. We will show that our reward function improves the convergence compared to the reward function proposed in the literature. Note that the authors in \cite{1} defined the state for discrete power levels and this proves our point.

\item\begin{equation}\label{eq.16} r_{t}^{i,n} = \begin{cases} e^{(-(C_{o}^{(n)} - \Gamma^{o})^2)} - e^{(-C_{i}^{(n)})} ,&  \sum_{n=0}^{N_{sub}} p_{t}^{i,n} \leq P_{max}^{f} \\ -3,&   \sum_{n=0}^{N_{sub}} p_{t}^{i,n} > P_{max}^{f}\end{cases}\end{equation}

    The reward function defined by equation (\ref{eq.14}) does not take into consideration the femtocell capacity. Thus, we define the above reward function with the rationale of maximizing the femtocell capacity while maintaining the macrocell capacity at $\Gamma^{o}$.
\end{enumerate}
\end{itemize}

\section{Performance Evaluation}\label{scenario}

\subsection{Simulation Scenario}

We consider a wireless network consisting of one macrocell underlaid with $N_{femto}$ femtocells. In the simulations, $N_{femto}$ ranges from $4$ to $15$ femtocells. Each femtocell serves $U_f = 1$ femto-user which is randomly located in the femtocell coverage area. Both the macro and femto cells share the same frequency band composed of $N_{sub} = 6$ subcarriers where orthogonal downlink transmission is assumed. The channel gain between transmitter $i$ and receiver $j$ on subcarrier $n$ is assumed to be path-loss dominated and is given by:

\begin{equation}\label{eq.11}
h_{ij}^{(n)} = d_{ij}^{(-k)}
\end{equation}

Where $d_{ij}$ is the physical distance between transmitter $i$ and receiver $j$, and $k$ is the path loss exponent. In the simulation $k = 2$ is used. The distances are calculated according to the following assumptions:

\begin{itemize}
\item The maximum distance between the MBS and its associated user is set to $1000$ meters.
\item The maximum distance between the MBS and a femto-user is set to $800$ meters.
\item The maximum distance between a FBS and its associated user is set to $80$ meters.
\item The maximum distance between a FBS and another femtocell's user is set to $300$ meters.
\item The maximum distance between a FBS and the macro-user is set to $800$ meters.
\end{itemize}

We used MatLab to simulate such scenario, where in the simulations we set the noise power $\sigma^2$ to $10^{-7}$, the maximum transmission power of the macrocell $P_{max}^{m}$ to $43$ dBm, the learning rate $\alpha$ to $0.5$, the discounted rate $\gamma$ to $0.9$ and the random number $\epsilon$ to $0.1$ during the first $80\%$ of the Q-iterations \cite{3}, \cite{4} and \cite{1}.

\subsection{Numerical Results}\label{results}

\begin{figure}[!t]
\centering
\includegraphics[height=0.45\columnwidth,width=1\columnwidth]{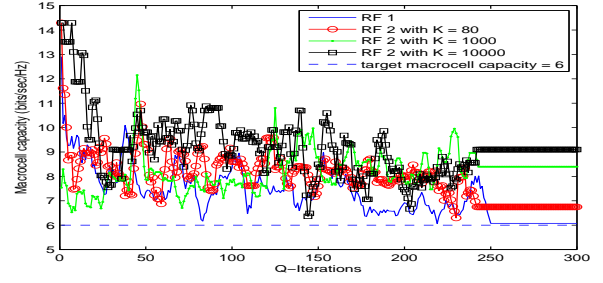}
\caption{Convergence of the macrocell capacity using different reward functions with $N_{femto} = 4$ with target capacity $=6$.}
\label{fig.2}
\end{figure}

We will refer to the reward functions defined by equations (\ref{eq.14}), (\ref{eq.15}) and (\ref{eq.16}) as $RF 1$, $RF 2$ and $RF 3$ respectively in all the simulations. Figure (\ref{fig.2}) shows the convergence of the macrocell capacity on a certain subcarrier ($C_{o}^{(n)}$) using $RF 1$ and $RF 2$ with $K=80$, $K=1000$ and $K=10000$. It can be observed that $RF 1$ shows better convergence behavior than $RF 2$ with both values of K (i.e: $RF 1$ converges to the target capacity ($\Gamma_{o} = 6$) more accurately). Moreover, the figure shows that the value of K affects the convergence where $K=80$ is better than $K=1000$ and $K=1000$ is better than $K=10000$, which proves our point that as the difference between the positive and negative rewards decreases, the convergence is enhanced. Note that in the simulations, the number of Q-iterations was $3000$ while in the figure only $300$ iterations are shown (i.e: The figure is drawn with step $= 10$) in order to achieve better resolution.

\begin{figure}[!t]
\centering
\includegraphics[height=0.45\columnwidth,width=1\columnwidth]{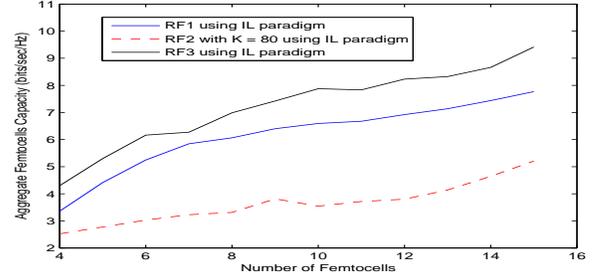}
\caption{Total femtocell capacity as a function of the number of femtocells using $RF 1$, $RF 2$ with $K=80$ and $RF 3$ in the IL paradigm.}
\label{fig.3}
\end{figure}

\begin{figure}[!t]
\centering
\includegraphics[height=0.45\columnwidth,width=1\columnwidth]{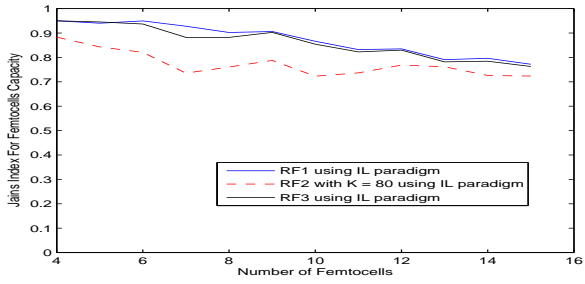}
\caption{Jain's fairness index (in terms of capacity) as a function of the number of femtocells $RF 1$, $RF 2$ with $K=80$ and $RF 3$ in the IL paradigm.}
\label{fig.5}
\end{figure}

Figure (\ref{fig.3}) shows the total femtocell capacity using $RF 1$, $RF 2$ with $K=80$ and $RF 3$ in the IL paradigm. It can be observed that introducing $C_{i}^{(n)}$ in $RF 3$ increases the aggregate femtocell capacity compared to $RF 1$. However, since the IL paradigm is used here, the femtocells act in a selfish way, which may reduce the fairness (in terms of capacity) between the femtocells compared to $RF 1$. This is shown in figure (\ref{fig.5}). Note that the fairness is evaluated using Jain's fairness index\cite{fair}: $f(x_1,x_2,\cdots,x_n) = \frac{(\sum_{i=1}^{n}x_{i})^2}{n\sum_{i=1}^{n}x_i^2}$ where $0 \leq f(x_1,x_2,\cdots,x_n) \leq 1$ and the equality to $1$ occurs when all the femtocells achieve the same capacity.

\begin{figure}[!t]
\centering
\includegraphics[height=0.45\columnwidth,width=1\columnwidth]{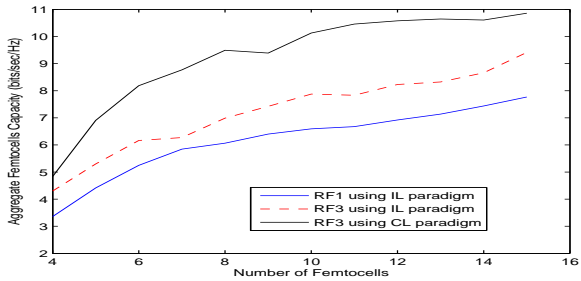}
\caption{Total femtocell capacity as a function of the number of femtocells using $RF1$ in the IL paradigm and $RF 3$ in both the IL and CL paradigms.}
\label{fig.7}
\end{figure}

\begin{figure}[!t]
\centering
\includegraphics[height=0.45\columnwidth,width=1\columnwidth]{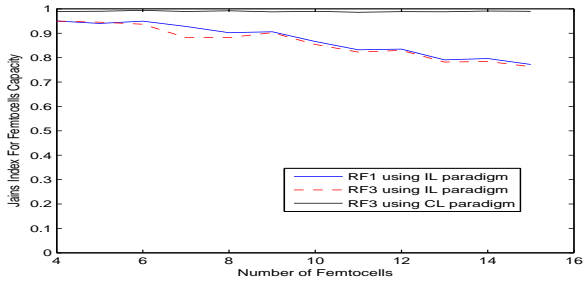}
\caption{Jain's fairness index (in terms of capacity) as a function of the number of femtocells using $RF1$ in the IL paradigm and $RF 3$ in both the IL and CL paradigms.}
\label{fig.9}
\end{figure}

\begin{figure}[!t]
\centering
\includegraphics[height=0.45\columnwidth,width=1\columnwidth]{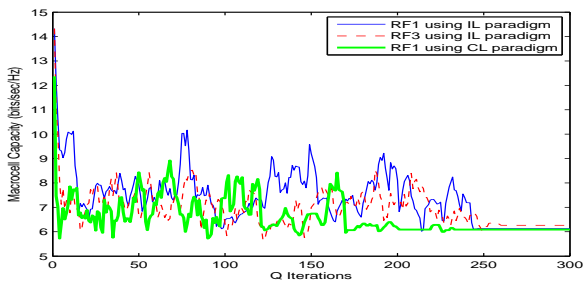}
\caption{Convergence of the macrocell capacity using $RF 1$ in the IL and CL paradigms and $RF 3$ in the IL paradigm with $N_{femto} = 4$ and target capacity $= 6$.}
\label{fig.11}
\end{figure}

As for the cooperation effect, figure (\ref{fig.7}) shows the total femtocell capacity using $RF 1$ in the IL paradigm and $RF 3$ in both IL and CL paradigms. From the figure, it can be noticed that introducing cooperation increases the total femtocell capacity. Actually, it can be observed that at $N_{femto} = 11$ cooperation increased the capacity by around $2.6$ bits/sec/Hz. Figure (\ref{fig.9}) shows that cooperation not only increases the capacity but also enhances the fairness. Moreover, figure (\ref{fig.11}) shows that cooperation speeds up the convergence (In the CL paradigm, convergence almost started after $1800$ iterations).

\section{Conclusion and Future work}\label{conclusion}

In this paper, a distributed Q-learning algorithm based on the multi-agent systems theory called DPC-Q is presented to perform power allocation in cognitive femtocells network. The DPC-Q algorithm is applied in two different paradigms: independent and cooperative. In the independent paradigm, two scenarios were considered. The first scenario is to control the interference generated by the femtocells on the macro-user where the results showed that the proposed algorithm is capable of maintaining the capacity of the macro-user at a certain threshold. The second scenario is to enhance the aggregate capacity of femtocells while maintaining the QoS of the macro-user. Through simulations, we showed that the independent learning paradigm can be used to increase the aggregate femtocell capacity. However, due to the selfishness of the femtocells, fairness is reduced compared to the first scenario. Thus, we proposed a cooperative paradigm, in which, femtocells share a portion of their Q-tables with each other. Simulation results showed that cooperation is capable of increasing the aggregate femtocell capacity and enhancing the fairness compared to the independent paradigm, with a relatively small overhead. Future works will focus on: $1)$ Devise a numerical framework to study the effect of changing the Q-learning parameters (i.e: $\gamma$, $\epsilon$ and $\alpha$) on the performance of the proposed algorithm $2)$ Design a control protocol to exchange the cooperation information amongst all femtocells $3)$ Other techniques for cooperation $4)$ Extending cooperation to coordination in which the femtocells try to coordinate their actions with each other to achieve the optimum joint action.

\section*{Acknowledgment}

This work is supported by the Qatar Telecom (Qtel) Grant No.QUEX-Qtel-09/10-10.

\bibliographystyle{./IEEEtran}
\bibliography{./bare_conf}


\end{document}